\documentclass[final]{cvpr}
\usepackage{times}
\usepackage{graphicx,xcolor}
\usepackage{amsmath,amssymb,amsfonts,bm,mathtools}
\usepackage{algorithm,algpseudocode}
\usepackage{booktabs,dcolumn,colortbl,adjustbox,multirow,setspace}
\usepackage[pagebackref=true,breaklinks=true,colorlinks,bookmarks=false]{hyperref}
\usepackage{cleveref}

\def\cvprPaperID{8619}
\def\confYear{CVPR 2021}
\title{%
    \attack\@:
    Piercing Through Adversarial Defenses
    with Latent Features%
}
\newcommand{\email}[1]{\href{mailto:#1}{\texttt{#1}}}
\newcommand*\samethanks[1][\value{footnote}]{\footnotemark[#1]}
\author{%
    Yunrui Yu\thanks{%
        These authors contributed equally to this work.
    } \\
    \hspace{-5pt}
    State key lab of IoTSC,
    \hspace{-5pt} \\
    University of Macau, \\
    Macau SAR, China.
    \\ {\small\email{yb97445@um.edu.mo}}
    \and
    Xitong Gao\samethanks[1] \\
    \hspace{-8pt}
    Shenzhen Institutes of Advanced Technology,
    \hspace{-8pt} \\
    Chinese Academy of Sciences, \\
    Shenzhen, China.
    \\ {\small\email{xt.gao@siat.ac.cn}}
    \and
    Cheng-Zhong Xu\thanks{
        Corresponding author.
    } \\
    State key lab of IoTSC, \\
    University of Macau, \\
    Macau SAR, China.
    \\ {\small\email{czxu@um.edu.mo}}
}

\graphicspath{{figures/}}
\DeclareRobustCommand{\inlinegraphics}[1]{%
    \begingroup\normalfont
    \includegraphics[height=\fontcharht\font`\B]{inline/#1}%
    \endgroup
}

\newcommand{\supdagger}{\textsuperscript{\textdagger}}
\newcommand{\supddagger}{\textsuperscript{\textdaggerdbl}}
\makeatletter%
\newcolumntype{L}{D{.}{.}{2,2}}
\newcolumntype{B}[3]{>{\boldmath\DC@{#1}{#2}{#3}}c<{\DC@end}}
\newcommand{\tbnum}[1]{\multicolumn{1}{B{.}{.}{2,2}}{#1}}
\newcommand{\tcenter}[1]{\multicolumn{1}{c}{#1}}
\newcommand{\thead}[1]{\multicolumn{1}{c}{\textbf{#1}}}

\makeatother%

\newcommand{\numero}[1]{No\@.\ {#1}}
\newcommand{\numeros}[1]{Nos\@.\ {#1}}

\newcommand{\unitk}{\,\text{k}}

\newcommand{\ordinal}[1]{{#1}\textsuperscript{th}}

\newcommand{\textsupsub}[2]{\( ^{\text{#1}}_{\text{#2}} \)}
\newcommand{\cifarx}{CIFAR-10}
\newcommand{\cifarc}{CIFAR-100}

\newcommand{\wrn}{WideResNet}

\newcommand\attack{{{LAFEAT}}}
\newcommand\att[2]{\textbf{LAF}\textsupsub{\,\texttt{#1}}{#2}}
\newcommand{\attmt}[1]{\att{MT}{#1}}
\newcommand{\apgddlr}{APGD\textsubscript{DLR}}

\DeclarePairedDelimiter\norm{\|}{\|}
\newcommand{\expect}[2]{\mathsf{E}_{#1}\left[{#2}\right]}
\newcommand{\textlnorm}[1]{\( \ell_{#1} \)-norm}
\newcommand{\lnorm}[2]{{\norm{#2}}_{#1}}
\newcommand{\trainset}{\mathcal{D}_\mathrm{train}}
\newcommand{\attackset}{\mathcal{D}_\mathrm{attack}}
\newcommand{\uniform}[1]{\mathcal{U}\left({#1}\right)}
\newcommand{\realset}{\mathbb{R}}
\newcommand{\idx}[2]{{#1}^{\scriptscriptstyle({#2})}}
\newcommand{\x}{\mathbf{x}}
\newcommand{\xx}[1]{\idx\x{#1}}
\newcommand{\xadv}{\hat\x}
\newcommand{\y}{\mathbf{y}}
\newcommand{\z}{\mathbf{z}}

\newcommand{\f}[1]{\idx{f}{#1}}
\newcommand{\m}{\bm{\mu}}
\newcommand{\logit}[1]{\idx{h}{#1}}
\newcommand{\logitweight}[1]{\idx{\lambda}{#1}}

\newcommand{\allw}{{\bm\theta}}
\newcommand{\target}{{\bm\tau}}
\newcommand{\inputset}{\mathcal{I}}
\newcommand{\intensorset}{%
    \ensuremath\realset^{{C}\times{H}\times{W}}}
\newcommand{\outputset}{\realset^K}
\newcommand{\stepsize}{\alpha}
\newcommand{\momentum}{\nu}
\newcommand{\dist}[2]{%
    \operatorname{d}\left({#1},{#2}\right)}
\DeclareMathOperator{\pool}{\mathrm{pool}}
\DeclareMathOperator{\sceloss}{%
    \mathcal{L}^\mathrm{sce}}
\DeclareMathOperator{\surrogateloss}{%
    \mathcal{L}^\mathrm{sur}}
\DeclareMathOperator{\targetloss}{%
    \mathcal{L}^\mathrm{sur}_\target}
\DeclareMathOperator{\latentloss}{%
    \mathcal{L}^\mathrm{lf}_{\bm\lambda}}
\DeclareMathOperator{\project}{%
    \mathcal{P}_{\epsilon, \x}}
\DeclareMathOperator{\sign}{\mathrm{sign}}
\DeclareMathOperator{\pgd}{\mathrm{PGD}_{\epsilon, \x, \y}}
\DeclareMathOperator{\onehot}{\mathrm{onehot}}

\begin{document}

\maketitle
\begin{abstract}
    Deep convolutional neural networks
    are susceptible to adversarial attacks.
    They can be easily deceived
    to give an incorrect output
    by adding a tiny perturbation to the input.
    This presents a great challenge
    in making CNNs robust against such attacks.
    An influx of new defense techniques
    have been proposed to this end.
    In this paper,
    we show that
    latent features in certain ``robust'' models
    are surprisingly susceptible
    to adversarial attacks.
    On top of this,
    we introduce
    a unified \textlnorm{\infty}
    white-box attack algorithm
    which harnesses latent features
    in its gradient descent steps,
    namely \attack.
    We show that not only
    is it computationally much more efficient
    for successful attacks,
    but it is also a stronger adversary
    than the current state-of-the-art
    across a wide range of defense mechanisms.
    This suggests that
    model robustness
    could be contingent
    on the effective use
    of the defender's hidden components,
    and it should no longer be viewed
    from a holistic perspective.
\end{abstract}

\section{Introduction}\label{sec:intro}

Many safety-critical systems,
such as
aviation~\cite{cheng16,akccay16,nips19drone}
medical diagnosis~\cite{wang18medical,li19clu},
self-driving~\cite{bojarski18,li19stereo,toromanoff20}
have seen a large-scale deployment
of deep \emph{convolutional neural networks} (CNNs).
Yet CNNs
are prone to \emph{adversarial attacks}:
a small specially-crafted perturbation
imperceptible to human,
when added to an input image,
could result in a drastic change
of the output of a CNN~\cite{
    szegedy14,goodfellow15,carlini19}.
As a rapidly increasing
number of safety-critical systems
are automated by CNNs,
it is now incumbent upon us
to make them robust
against adversarial attacks.

The strongest assumption commonly used
for generating adversarial inputs
is known as \emph{white-box attacks},
where the adversary
have full knowledge
of the model~\cite{carlini17}.
For instance,
the model architecture, parameters,
and training algorithm and dataset
are completely exposed to the attacker.
By leveraging the gradient of the output loss
\emph{with respect to} (\wrt) the input,
gradient-based methods~\cite{
    moosavi2016deepfool,carlini17,madry18}
have been shown
to decimate the accuracy of CNNs
when evaluated on adversarial examples.
Many new techniques
to improve the robustness of CNNs
have since been proposed
to defend against such attacks.
Recent years
have therefore seen a tug of war
between adversarial attack~\cite{
    goodfellow15,moosavi2016deepfool,carlini17,
    madry18,dong18momentum,xiao2019advgan,croce20aa}
and defense~\cite{
    madry18,shafahi19free,carmon19,alayrac19,zhang19trades,
    pang20hypersphere,wang20misclass,wu20wp,wu20width,gowal20}
strategies.
Attackers search for a perturbation
that maximizes the loss of the model output,
typically through gradient ascent methods,
\eg~one popular method
is \emph{projected gradient descent} (PGD)~\cite{madry18};
whereas defenders attempts
to make the loss landscape smoother
\wrt{} the perturbation
via adversarial training,
\ie~training with adversarial examples.

\begin{figure}[ht]
    \centering%
    \newcommand{\mags}[1]{%
        \includegraphics[
            width=0.32\linewidth, trim=0 10pt 40pt 30pt
        ]{layers/mags#1}}
    \mags{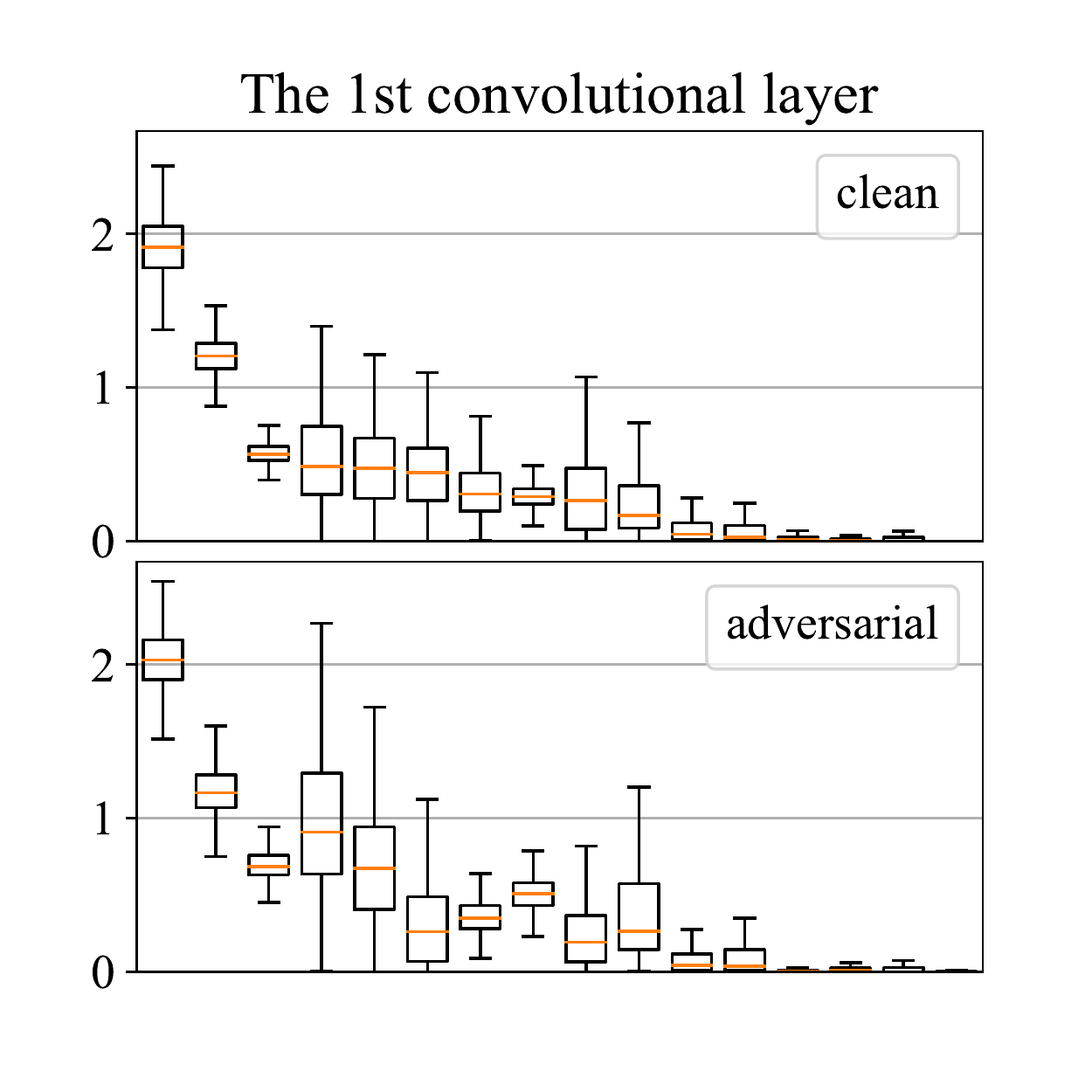} \mags{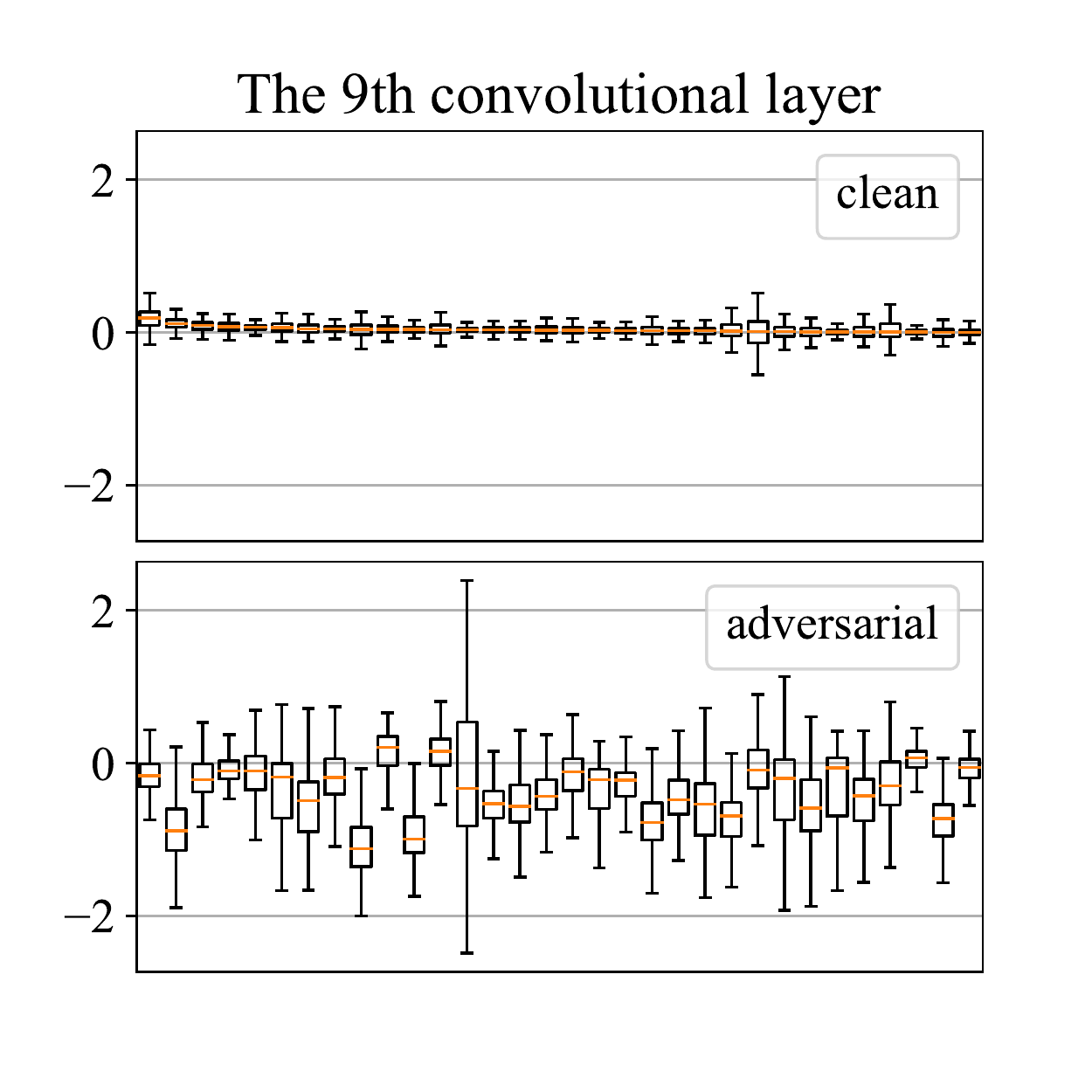} \mags{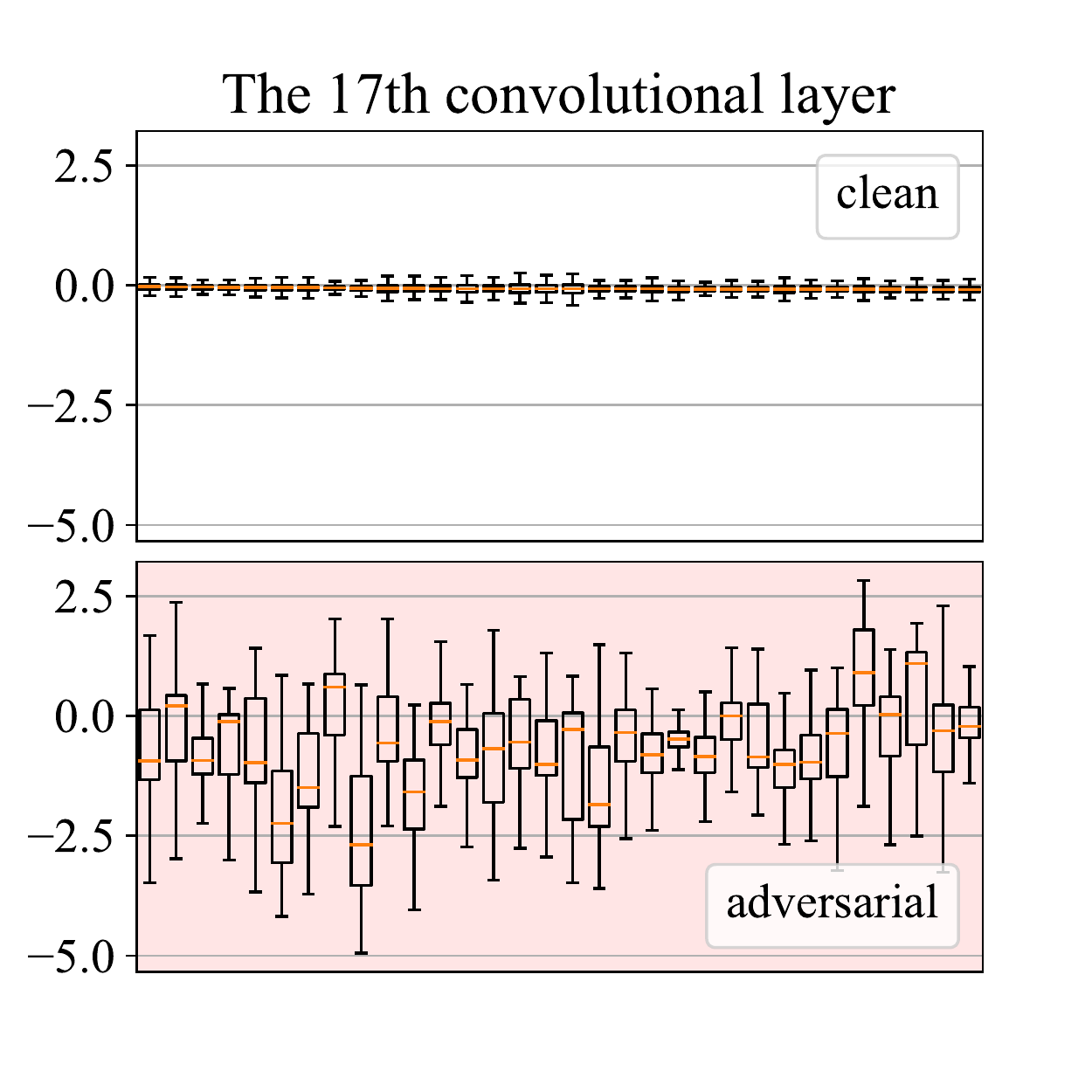}
    \mags{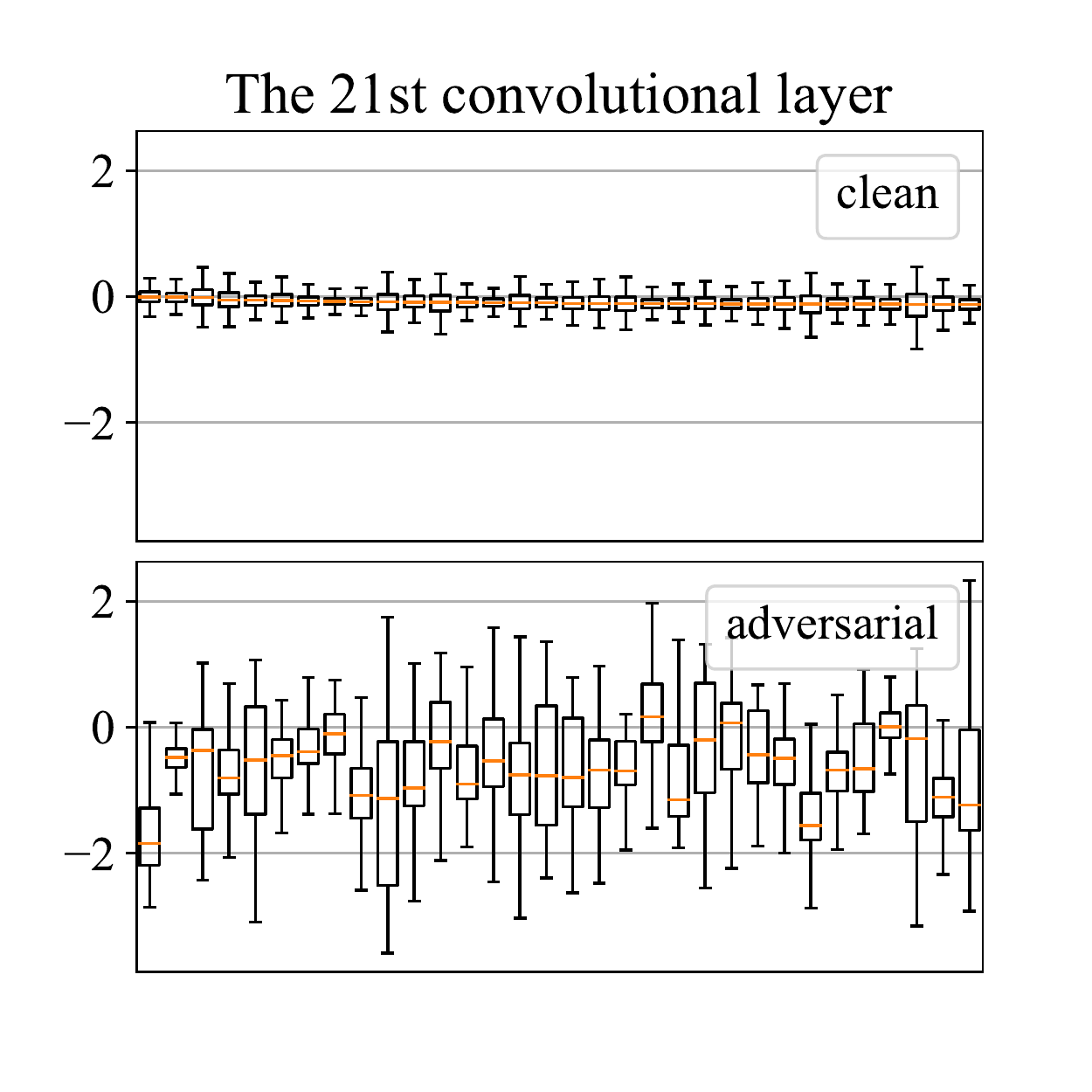} \mags{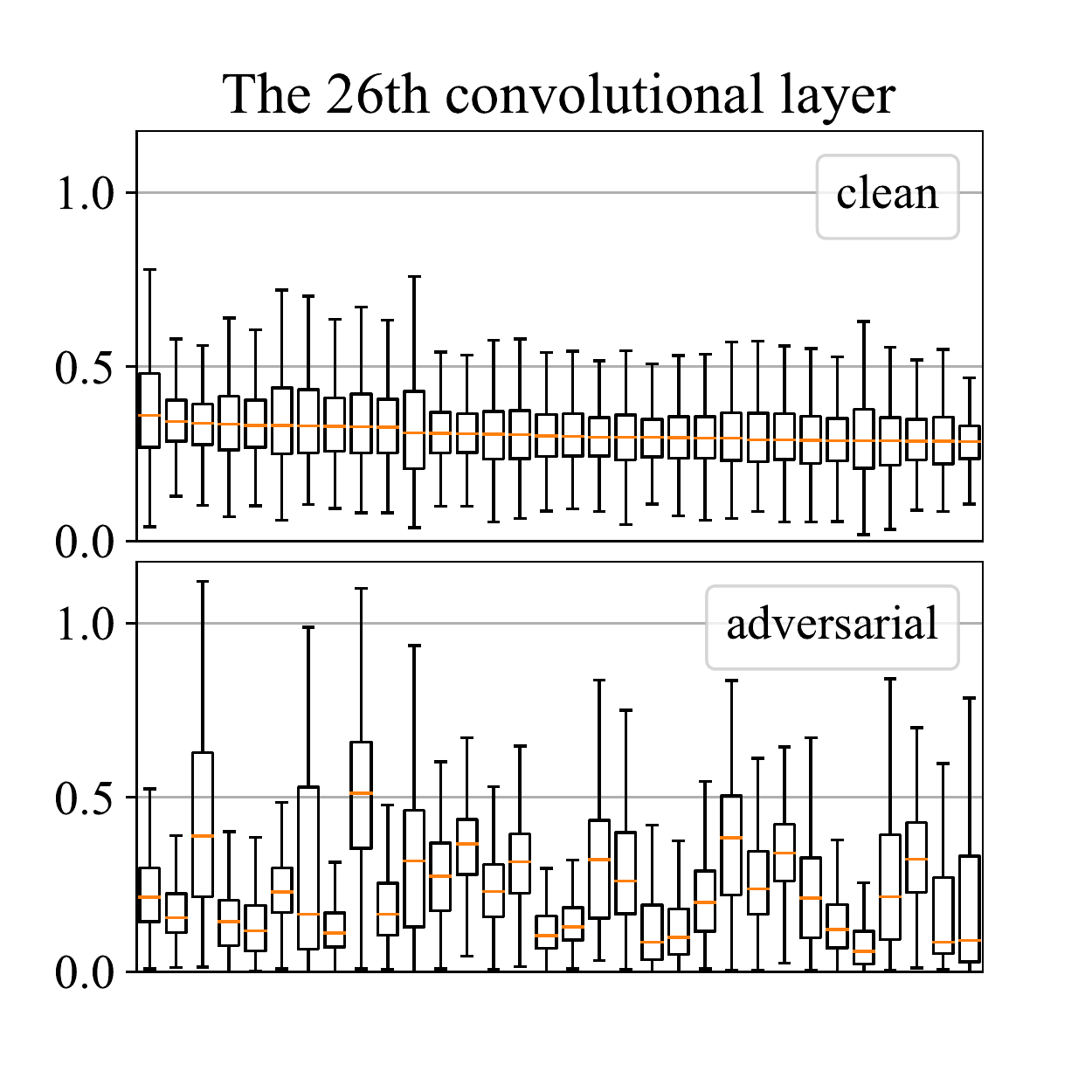} \mags{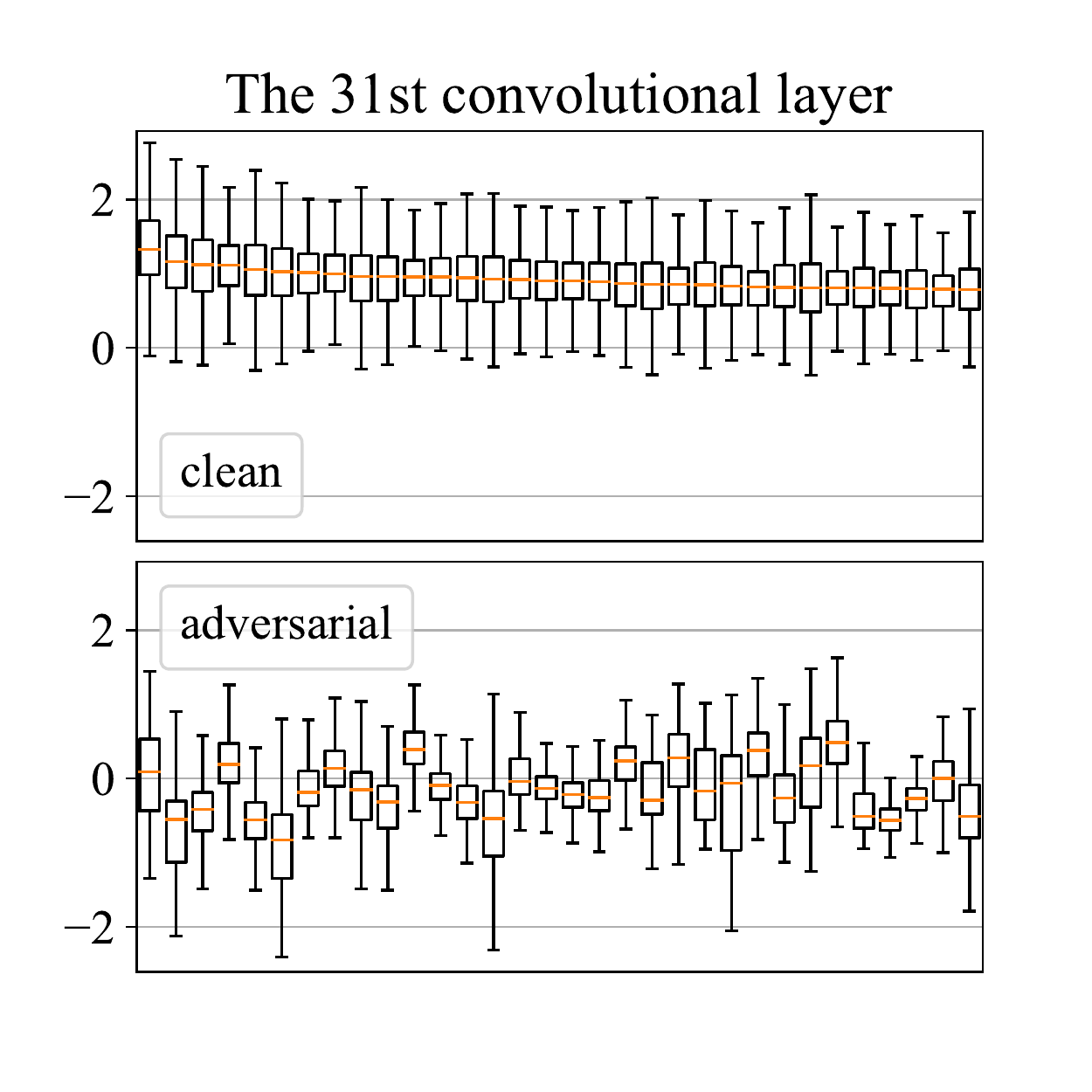}
    \caption{%
        The distribution
        of 32 most excited channels on average
        of convolutional layers
        in a naturally trained \wrn-32-10
        when being shown \cifarx{} ``airplane'' images.
        We compare that against
        the corresponding channel activations
        under adversarial ``airplane'' examples
        with LPGD-10.
        Attacking the \ordinal{17} layer
        (shaded in red)
        resulted in scrambled features
        across the entire model
        and incorrect final model outputs.
    }\label{fig:intro:mags}
\end{figure}

From a human perception perspective,
as feature extractors,
shallow layers of CNNs
extract simple local textures
while neurons in deep layers
specialize to differentiate
complex objects~\cite{
    olah17feature,carter19activation}.
Intuitively, we expect
incorrectly extracted shallow features
often cannot be pieced together
to form correct high-level features.
Moreover,
this could have a cascading effect
in subsequent layers.
To illustrate,
we equipped PGD with the ability
to attack \emph{one}
of the intermediate layers
by maximizing \emph{only} the loss
of an attacker-trained classifier,
which we call LPGD for now.
In~\Cref{fig:intro:mags},
we scrambled the feature extracted
by attacking an intermediate layer with LPGD\@,
and observed
increasing discrepancies
between the pairs of features extracted
from the natural images
and their associated adversary
in deeper layers.

Nevertheless,
existing attack and defense strategies
approach the challenge
of evaluating or promoting
the white-box model robustness
in a \emph{model-holistic} manner.
Namely, for classifiers,
they regard the model
as a single non-linear differentiable function \( f \)
that maps the input image
to output logits.
While these approaches
generalize well across models,
they tend to ignore
the latent features extracted
by the intermediate layers within the model.

Some recent defense strategies~\cite{%
    atzmon19levelsets,zhang19scatter,zhang20interpolation,
    kim20sensible,jin20manifold,mustafa19pcl,
    naseer20stylized}
reported that their models
can achieve high robustness
against PGD attacks.
Understandably,
these defenses
are highly specialized
to counter these conventional attacks.
We speculate
that one of the reasons
why PGD failed to break through the defenses
is because of its \emph{model-holistic} nature.
This notion implores us to ask two important questions:
    \emph{%
        Can latent features be vulnerable
        to attacks;
        and subsequently,
        can the falsely extracted features
        be cascaded to the remaining layers
        to make the model output incorrect?
    }

\begin{figure}[ht]
    \centering%
    \includegraphics[width=\linewidth, trim=0 10pt 0 0]{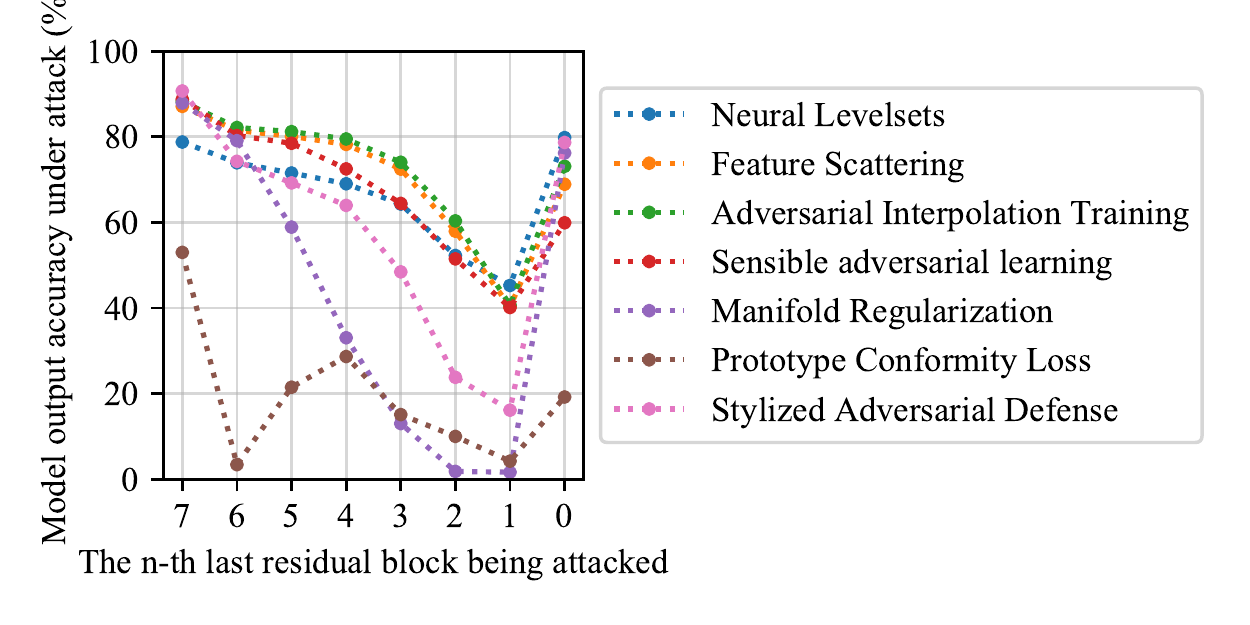}
    \caption{%
        We used LPGD-100,
        or PGD with 100 iterations
        to attack
        \emph{only} the \ordinal{\(n\)}
        residual block
        of a ``robust'' model.
        Except the \ordinal{0}
        denotes the output layer
        and PGD is used\@.
        To our surprise,
        adversaries generated
        by attacking an \emph{early} layer
        are highly transferable
        to the \emph{final} classification output
        of the model.
        Defending models
        are respectively obtained from~\cite{
            atzmon19levelsets,zhang19scatter,
            zhang20interpolation,kim20sensible,
            jin20manifold,mustafa19pcl,huang20sat}.
    }\label{fig:intro:attack}
\end{figure}

It turns out that
the new adversarial examples
computed by LPGD
can harm the accuracies
of the ``robust'' models above
(\Cref{fig:intro:attack}).
The experiment showed that
while they are trained
to be effective against PGD,
they could fail spectacularly
when faced attacks
that simply target their latent features.
This may also imply that
a flat model loss landscape
\wrt{} the input image
does not necessarily entail
flat latent features \wrt{} the input.
Existing attack methods
that rely on a holistic view of the model
therefore may fail
to provide a reliable assessment
of model robustness.

Motivated by the findings above,
in this paper
we propose a new strategy,
\attack,
which seeks to harness latent features
in a generalized framework.
To push the envelope
of current state-of-the-art (SOTA)
in adversarial robustness assessment,
it draws inspiration
from effective techniques
discovered in recent years,
such as the use of momentum~\cite{dong18momentum,croce20aa},
surrogate loss~\cite{gowal20,ding20mma},
step size schedule~\cite{croce20aa,gowal19surrogate},
and multi-targeted attacks~\cite{
    gowal19surrogate,qin19ll,pang20hypersphere}.
To summarize,
our main contributions
are as follows:
\begin{itemize}
    \item We introduce
    how intermediate layers
    can be leveraged
    in adversarial attacks.

    \item We show that
    latent features
    provide faster convergence,
    and accelerate gradient-based attacks.

    \item By combining multiple
    effective attack tactics,
    we propose \attack.
    Empirical results
    show that it rivals competing methods
    in both the attack performance
    and computational efficiency.
    We perform extensive ablation analysis
    of its hyperparameters
    and components.
\end{itemize}

To the best of our knowledge,
\attack{}
is currently the strongest
against a wide variety of defense mechanisms
and matches the current top-1
on the TRADES~\cite{zhang19trades}
\cifarx{} white-box leaderboard (\Cref{sec:results}).
Since latent features
are vulnerable to adversarial attacks,
which could in turn break robust models,
we believe the future evaluation
of model robustness
could be contingent
on how to make effective use
of the hidden components
of a defending model.
In short,
model robustness
should no longer be viewed
from a holistic perspective.

\section{%
    Preliminaries \& Related Work%
}\label{sec:related}

\subsection{Adversarial examples}\label{sec:related:attack}

Before one can
generate adversarial examples
and defend against them,
it is necessary
to formalize the notion
of \emph{adversarial examples}.
We start by defining
the classifier model
as a function
\( f_\allw ( \x ) \),
where \( \allw \)
is the model parameters,
\( f_\allw \colon \inputset \to \outputset \)
maps the input image
to its classification result,
\( \inputset \subset \intensorset \)
limits the image data to a valid range,
with \( C \) being the number of channels
in the image
(typically 3 channels for a colored input),
\( H \) and \( W \) the height and width
of the image respectively,
and \( K \) is the number of classes
from the model output.

The attacker's objective
is to find an \emph{adversarial example}
\( \xadv \in \inputset \)
of the model under attack \( f_\allw \)
by (approximately) solving the optimization problem:
\begin{equation}
    \max_{\mathclap{
        \xadv \in \inputset \wedge
        \dist\x\xadv \leq \epsilon
    }}
    \,\sceloss\left(
        f_\allw\left( \xadv \right), \y
    \right),
    \label{eq:adversarial}
\end{equation}
where \(
    \sceloss\left(
        f_\allw\left( \xadv \right), \y
    \right)
\)
is the \emph{softmax cross-entropy} (SCE) loss
between the output
and the one-hot ground truth \( \y \).
By maximizing the loss,
one may arrive at a \( \xadv \)
such that \( \arg\max f_\allw(\xadv) \neq \arg\max \y \).
In other words,
the model
can generally be fooled
to produce an incorrect classification.
To confine the perturbation,
\( \dist{\x}{\xadv} \leq \epsilon \)
constrains the distance
between the original \( \x \)
and the adversarial \( \xadv \)
to be less than or equal some small constant
\( \epsilon \).

In general,
the distance metric \( \dist{\x}{\xadv} \)
is commonly defined
as the \textlnorm{p} of the difference
between \( \x \) and \( \xadv \)~\cite{
    szegedy14,goodfellow15,madry18,carlini19}.
Different choices of norm
were explored in literature,
\eg~one pixel attacks~\cite{su17onepixel}
minimizes the \textlnorm{0}
\( {\lnorm{0}{\x - \xadv}} \),
while others may be interested
in the standard Euclidean distance,
the \textlnorm{2}~\cite{
    szegedy14,moosavi2016deepfool,carlini17}.
In this paper,
we focus on another popular choice of distance metric,
the \textlnorm{\infty}
\(
    \dist{\x}{\xadv}
    \triangleq
    \lnorm{\infty}{\x - \xadv}
\),
as used in~\cite{goodfellow15,madry18}.

The white-box scenario
completely exposes to the attacker
to the inner mechanisms of the defense,
\ie~the model architecture, its parameters,
training data and algorithms, and \etc{}
are revealed to the attacker.
The optimal solution
of \eqref{eq:adversarial},
however, is generally unattainable.
In practice, approximate solution
are instead sought after,
often with gradient-based methods,
\eg~one of the popular attack method
used by defenders
for evaluating white-box adversarial robustness
is the \emph{projected gradient descent} (PGD).
PGD finds an adversarial example
by performing the iterative update~\cite{madry18}:
\begin{equation}
    \xadv_{i+1} = \project \left(
        \xadv_{i} + \alpha_i \sign\left(
            \nabla_{\xadv_i} \sceloss\left(
                f_\allw(\xadv_i), \y
            \right)
        \right)
    \right).
    \label{eq:pgd}
\end{equation}
Initially,
\( \xadv_0 = \project (\x + \mathbf{u}) \),
where \( \mathbf{u} \sim \uniform{[-\epsilon, \epsilon]} \),
\ie~\( \mathbf{u} \) is a uniform random noise
bounded by \([-\epsilon, \epsilon]\).
The function
\( \project \colon \intensorset \to \inputset \)
clips the range of its input
into the \(\epsilon\)-ball neighbor
and the \( \inputset \).
The term \(
    \nabla_{\xadv_i} \sceloss\left(
        f_\allw(\xadv_i), \y
    \right)
\) computes the gradient of the loss
\wrt{} the input \( \xadv_i \).
Finally,
\( \alpha_i \) is the step size,
and for each element in the tensor \( \z \),
\( \sign(\z) \) returns
one of \( 1 \), \( 0 \) or \( -1 \),
if the value is positive, zero or negative respectively.
For simplicity, we define:
\begin{equation}
    \pgd(\mathcal{L}, \bm\stepsize, i) \triangleq \xadv_i,
    \label{eq:pgd:def}
\end{equation}
\ie~the result of iterating
for \( i \) times
with a sequence of step sizes \( \bm\stepsize \)
and the loss function \( \mathcal{L} \)
on the original image \( \x \).
Other gradient-based methods
include fast gradient-sign method (FGSM)~\cite{goodfellow15},
basic iterative method (BIM)~\cite{kurakin17adversarial},
momentum iterative method (MIM)~\cite{dong18momentum}
and fast adaptive boundary attack (FAB)~\cite{croce20fab}
for \textlnorm{\infty} attacks,
Carlini and Wagner (C\&W)~\cite{carlini17}
for \textlnorm{2} attacks,
and DeepFool~\cite{moosavi2016deepfool} for both.
Similar to PGD,
they only iteratively
leverage the loss gradient \(
    \nabla_{\xadv_i} \mathcal{L}\left( f_\allw(\xadv_i), \cdots \right)
\),
as all of them adopt a holistic view
on the model \( f_\allw \).

Because the SCE loss function
\( \sceloss \)
in the objective~\eqref{eq:adversarial}
is highly non-linear,
easily saturated,
and normally evaluated
with limited floating-point precision,
gradient-based attacks
may experience vanishing gradients
and difficulty converging~\cite{carlini17,croce20aa}.
Recent attack methods
hence use \emph{surrogate losses} instead
for gradient calculation~\cite{
    carlini17,gowal19surrogate},
and optimize an alternative objective
by replacing \( \sceloss \)
with a custom surrogate loss function.
As the alternative objective
is usually aligned with the original,
maximizing the latter
would also maximize the former.

Many auxiliary tricks
can push the limit of existing attack methods,
for instance,
a step-size schedule~\cite{croce20aa,gowal19surrogate}
with a decaying step-size
in relation to the iteration count
could improve the overall success rate.
Multi-targeted attack~\cite{
    gowal19surrogate,qin19ll,pang20hypersphere}
uses label-specific surrogate loss
by enumerating all possible target labels.
Attackers may also resort
to an ensemble of multiple attack strategies,
making the compound approach stronger
than any individual attacks~\cite{carlini19,croce20aa}.
The latter two methods, however,
tend to introduce an order of magnitude increase
in the worse-case computational costs.

Generative networks
that learn from the loss
were proposed
for adversarial example synthesis~\cite{
    baluja18atn,jang19generate}.
This tactic can be further enhanced with
generative adversarial networks (GANs)~\cite{goodfellow14gan},
where the discriminator network
encourages the distribution of adversarial examples
to become indistinguishable
from that of natural examples~\cite{
    xiao2019advgan,mangla2019advganpp}.

Finally,
there are a few recent publications
that leverages latent features
in their attacks~\cite{kumari19harnessing,khrulkov18}.
Unlike these methods,
\attack{} considers
\textlnorm{\infty}
white-box attacks,
and further differentiate itself from them
as it learns to attack defending models.

\subsection{Defending against adversarial examples}

The objective of robustness against
adversarial examples
can be formalized
as a saddle point problem,
which finds model parameters
that minimize the adversarial loss~\cite{madry18}:
\begin{equation}
    \min_\allw \expect{(\x, \y) \sim \trainset}{
        \qquad\!
        \max_{\mathclap{
            \xadv \in \inputset \wedge
            \dist\x\xadv \leq \epsilon
        }}
        \,\sceloss\left(
            f_\allw\left( \xadv \right), \y
        \right)
    },
    \label{eq:robustness}
\end{equation}
where \( \trainset \)
contains pairs of input images \( \x \)
and ground truth labels \( \y \).
A straightforward approach
to approximately solving
the above objective~\eqref{eq:robustness}
is via \emph{adversarial training}~\cite{kurakin17atscale},
which trains the model with adversarial examples
computed on-the-fly using,
for instance, PGD~\cite{madry18}.
\begin{figure*}[!ht]
    \centering%
    \includegraphics[width=\textwidth]{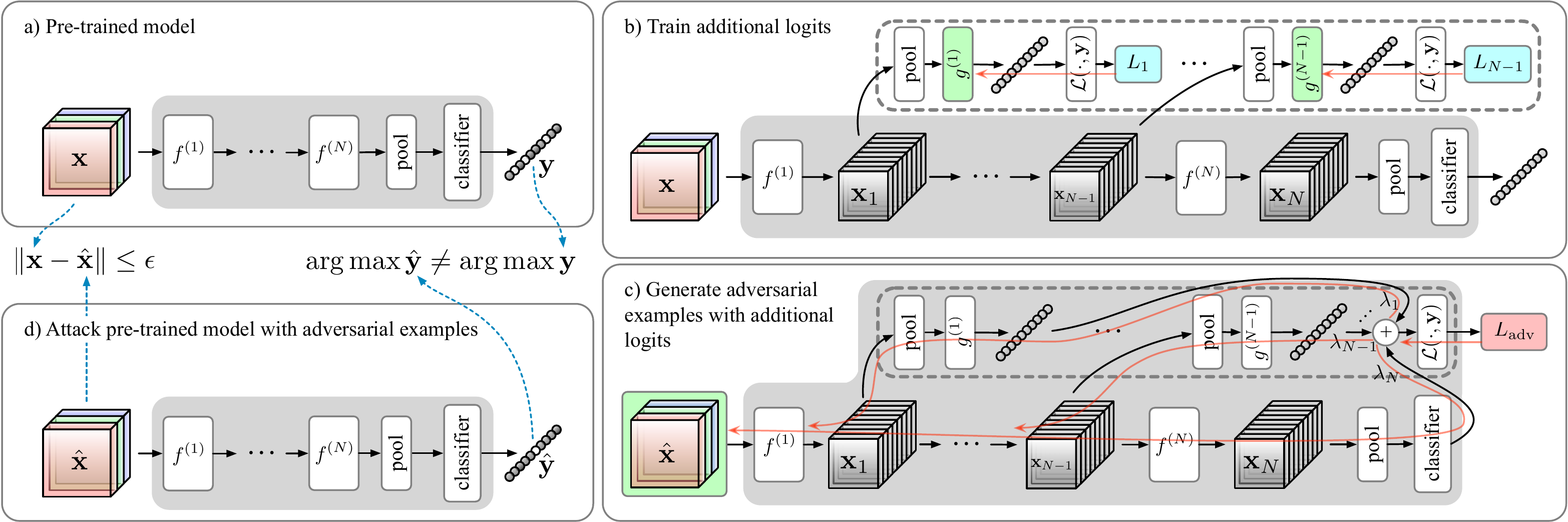}%
    \caption{%
        High-level overview of \attack.
        Note that in each of the steps,
        layers in \inlinegraphics{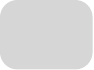} (gray) regions
        remain fixed,
        \inlinegraphics{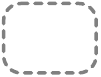} (dotted outline) regions
        denote new layers added by \attack,
        \inlinegraphics{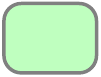} (green) layers
        are being trained,
        and losses in \inlinegraphics{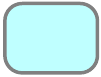} (cyan)
        and \inlinegraphics{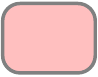} (red) blocks
        are respectively minimized and maximized.
        \textbf{(a)}
        illustrates the original pre-trained model,
        where \( \f{l} \)
        denotes the \ordinal{\(l\)} layer
        (or residual block).
        \textbf{(b)}
        trains additional fully-connected layers
        from intermediate layers,
        each with a softmax cross-entropy loss
        until convergence.
        \textbf{(c)}
        computes adversarial examples iteratively
        with a weighted sum of surrogate losses.
        \textbf{(d)}
        uses the adversarial examples
        from (c) to evaluate the robustness
        of the original model.
    }\label{fig:method:overview}
\end{figure*}

Many adversarial defense strategies
follow the same paradigm,
but train the model
with different loss objective functions
in order to further foster robustness.
Along with the standard classification loss,
TRADES~\cite{zhang19trades}
minimizes the multi-class calibrated loss
between the output of the original image
and the one of the adversarial example.
Misclassification-aware regularization~\cite{wang20misclass}
encourages the smoothness
of the network output,
even when it produces misclassified results.
Self-adaptive training~\cite{huang20selfadaptive}
allows the training algorithm
to adapt to noise added to the training data.
Feature scattering~\cite{zhang19scatter}
generates adversarial examples for training
by maximizing the distances
between features extracted
from the natural and adversarial examples.
Neural level-sets~\cite{atzmon19levelsets}
and sensible adversarial training~\cite{kim20sensible}
use different proxy robustness objectives
for adversarial training.
Hypersphere embedding~\cite{pang20hypersphere}
normalizes weights and features
to be on the surface of hyperspheres,
and also normalizes the angular margin
of the logits layer.
Prototype conformity loss~\cite{mustafa19pcl},
and manifold regularization~\cite{jin20manifold}
adopt different regularization losses
to allow the model
to learn a smooth loss landscape
\wrt{} changes in \( \x \).
Stylized adversarial defense~\cite{naseer20stylized}
and learning-to-learn (L2L)~\cite{jang19generate}
propose to use neural networks
to generate adversarial examples for training.
Finally, self-training
with unlabeled data~\cite{carmon19,alayrac19}
can substantially improve robustness
in a way that cannot be trivially broken
by adversarial attacks.

As robust models may be substantially
larger than non-robust ones,
there have been a recent interest
in making them more space and time efficient.
Alternating direction method of multipliers (ADMM)
has been applied
to prune and adversarial train CNNs
jointly~\cite{ye19compression}.
HYDRA~\cite{sehwag20hydra}
preserves the robustness of pruned models
by integrating the pruning objective
into the adversarial loss optimization.

Adversarial training
often requires several iterations
to compute adversarial examples
for each model parameter update,
which is multiple times more expensive
than traditional training
with natural examples.
Adversarial training for free~\cite{shafahi19free}
address this problem
by interleaving adversary updates
with model updates
for efficient training of robust CNNs.
Fast adversarial training~\cite{wong20fast}
further accelerates the training
by using a simpler FGSM-based adversary.


Finally, others provide
practical considerations and tricks
for stronger adversarial defense~\cite{
    carlini19,pang20bag,rice20overfitting,
    wu20width,chen20cvpr}.

\section{The \attack{} Method}\label{sec:method}


We introduce \attack{}
by providing a high-level illustration
(\Cref{fig:method:overview})
of its attack procedure.
First,
for a firm grip
on latent features,
it starts by training
fully-connected layers
for each residual block
with the training set
until convergence.
Note that we ensure
the original model \( f_\allw \)
to remain constant during this process.
To compute adversarial examples,
we maximize the alternative adversarial loss
\( \hat{L} \),
which is an adaptively-weighted sum
of surrogate losses from individual layers.
For testing of adversarial robustness,
the generated adversarial example
is then transferred
to the original model \( f_\allw \)
for evaluation.

\subsection{%
    Latent feature adversarial%
}\label{sec:method:lfa}

Following the footsteps
of surrogate losses,
in \Cref{sec:intro}
we postulate that
a similarly indirect loss on latent features
can also effectively enhance adversarial attacks.
\attack{}
exploits the features extracted
from intermediate layers
to craft even stronger adversarial examples
for \( f_\allw \).
We assume
the model architecture \( f_\allw \)
to generally comprise
a sequence of \( N \) layers
(or residual blocks)
and can be represented as:
\begin{equation}
    f_\allw (\x) = \f{N}( \cdots \f{2}( \f{1}(\x) ) \cdots ),
\end{equation}
where \( \f{1}, \f{2}, \ldots, \f{N} \)
denotes the sequence
of intermediate layers in the model.
For simplicity in notation,
we omit the parameters from individual layers.
We therefore
formalize this proposal by generalizing
the traditional PGD attack~\eqref{eq:pgd}
with a \emph{latent-feature PGD} (LFPGD) adversarial
optimization problem:
\begin{equation}
    \begin{aligned}
    & \max_{
        \mathclap{
            \bm{h},\,
            \bm\lambda,\,
            \bm\stepsize,\,
            \surrogateloss
        }
    }
    \quad\,
    \sceloss\!\left(
        f_\allw\left(
            \pgd\left( \latentloss, \bm\stepsize, I \right)
        \right), \y
    \right), \\
    \textit{where} \quad
    & \latentloss(\z) = \surrogateloss\!\left(
        {\textstyle\sum_{l \in [1:N]}}
        \logitweight{l} \logit{l} \left(
            \idx\z{l}
        \right), \y
    \right).
    \end{aligned}\label{eq:latent}
\end{equation}
Here, the constant \( I \) is the maximum number
of gradient-update iterations.
For each layer \( l \in {[1 : N]} \),
\( \idx\lambda{l} \in [0, 1] \)
assigns an importance value
to the layer gradient
with \( \sum_{l \in N} \idx\lambda{l} = 1 \).
The term
\(
    \idx\z{l}
    \triangleq
    \f{l} \circ \cdots \circ \f{1}(\z) \)
denotes the feature extracted
from the \ordinal{\(l\)} layer,
The function \( \logit{l} \)
for the \ordinal{\( l \)} layer
maps the features from \( \f{l} \) to logits.
Finally, \( \alpha_i \) is a step-size schedule.
Our goal is hence
to find the right combinations
of logits functions
\(
    \bm{h} \triangleq
    (\logit{1}, \ldots, \logit{N})
\)
and their corresponding weights
\(
    \bm\lambda \triangleq
    (\logitweight{1}, \ldots, \logitweight{N})
\)
from intermediate layers,
the step-size schedule \( \bm\alpha \),
and the surrogate loss \( \surrogateloss \) to use.

Solving the LFPGD optimization
is unfortunately infeasible in practice.
For which we devise methods
that could \emph{approximately} solve it,
and nevertheless enable us
to generate adversarial examples
stronger than competing methods.

\subsection{%
    Training intermediate logits layers%
}\label{sec:method:logits}

To utilize latent features,
we begin by training logits layers \( \logit{l} \)
for individual intermediate layers \( \f{l} \)
for all \( l \in [1:N - 1] \)
with conventional stochastic gradient descent (SGD)
until convergence.
The function \(
    \logit{l}\colon
    \realset^{
        \idx{C}{l} \times \idx{H}{l} \times \idx{W}{l}
    }
    \to \realset^K
\) is defined
as a small auxiliary classifier
composed of
a global average pooling layer
\(
    \pool\colon
    \realset^{
        \idx{C}{l} \times \idx{H}{l} \times \idx{W}{l}
    }
    \to \realset^{\idx{C}{l}}
\)
followed by a fully-connected layer for classification:
\begin{equation}
    \logit{l}(\xx{l}) \triangleq \pool\left(
        \xx{l}
    \right) \idx{\bm\phi}{l} + \idx{\bm\eta}{l},
\end{equation}
where
\( \xx{l} \)
denotes the features extracted
from the \ordinal{\(l\)} layer,
\(
    \idx{\bm\phi}{l} \in \realset^{\idx{C}{l} \times K}
\) and \(
    \idx{\bm\eta}{l} \in \realset^{K}
\) are parameters to be trained
in the function \( \logit{l} \).
As the final layer \( \f{N} \)
is already a logits layer,
we assume \( \logit{N} \triangleq \mathrm{id} \)
is an identity function.

Depending on the availability,
we could train the added layers
with either \( \trainset \),
the data samples used for attack \( \attackset \),
or both together.
While we used \( \trainset \)
in our experiments,
we observed in practice
negligible differences
in either attack strengths
given sufficient amount of training examples,
as they are theoretically drawn
from the same data sampling distribution.

It is important to note that
during the training procedure of \( \logit{l} \),
the original model \( f_\allw \)
is used as a feature extractor,
with all training techniques
(\eg~dropout, parameter update, \etc)
disabled.
This means that the model parameters \( \allw \),
the layers \( f_{l} \),
and their parameters,
batch normalization~\cite{ioffe15bn} statistics,
and \etc{}
remain \emph{constant},
while only the parameters
in \( \logit{l} \) functions
are being trained.

\subsection{%
    Choosing intermediate layers to attack%
}\label{sec:method:weight}

The search space for \( \logitweight{1:N} \)
is difficult to navigate
because of the computations cost
associated with finding a statistically significant
amount of adversarial examples.
For this reason,
\attack{}
simplifies the search
with a greedy yet effective process.
First, we enumerate
over all intermediate layers
\( l \in [1:N-1] \),
and let
\begin{equation}
    \latentloss(\z, \y) = \surrogateloss\left(
        {\textstyle
            \beta \logit{l}\!\left(
                \idx\z{l}
            \right) + (1 - \beta) f_\allw(\z), \y
        }
    \right)
\end{equation}
by setting \(
    \bm\lambda =
        \beta \onehot(N, N) + (1 - \beta)\onehot(l, N),
\)
where initially \( \beta = \frac12 \).
In other words,
the attack is now using only the \ordinal{\(l\)} layer
together with the output layer
at a time
while disabling all others.
With this method,
we can discover
the most effective layer
for subsequent attack procedures
across all images
in \( \attackset \).
Empirically,
we found in most defending models
the weakest link
is the penultimate residual block,
and this search procedure
can thus be skipped entirely
for performance considerations.
There are, however, exceptions:
for instance,
we found in the model
from~\cite{mustafa19pcl},
the \ordinal{6} last residual block
exhibits the weakest defense
and attacks using it converge faster.

Finally,
empirical results revealed that
the intermediate layer \( l \)
can be adaptively disabled
if it misclassifies the adversarial example,
\ie~when \( \arg\max\logit{l}(\xadv_i) \neq \y \),
in order to optimize
faster towards the original
adversarial objective~\eqref{eq:adversarial}.
We incorporate this in the final algorithm.

\subsection{%
    Surrogate loss function%
}\label{sec:method:surrogate}

Gradients computed
from the SCE loss
has been notoriously shown
to easily underflow
in floating-point arithmetic~\cite{
    carlini17,croce20aa}.
For this reason,
\emph{surrogate loss} functions
have been proposed~\cite{
    carlini17,gowal19surrogate,croce20aa}
to work around this limitation.
Despite their effectiveness
in breaking through defenses,
it is difficult
to interpret why they work
as they are no longer
maximize the original
adversarial objective~\eqref{eq:adversarial}
directly.
As a result,
we propose a small modification
to the original SCE loss,
which scales the logits adaptively
before evaluating the softmax operation:
\begin{equation}
    \surrogateloss(\z, \y) \triangleq \sceloss\!\left(
        \frac\z{t} {\left(
            \y^\top \z -
            \max\left(
                \left( 1 - \y \right) \cdot \z
            \right)
        \right)}^{-1}, \y
    \right),
    \label{eq:surrogate}
\end{equation}
where \( \cdot \) denotes the element-wise product,
and the temperature \( t = 1 \)
in all of our experiments.
Here, \(
    \y^\top \z -
    \max\left(
        {\left( 1 - \y \right)} \cdot \z
    \right)
\) is known as
the difference of logits (DL)~\cite{carlini17},
which evaluates the difference
between the largest output in the logits \( \z \)
against the second largest.
The negated version of DL
and the ratio-based variant---%
the difference of the logits ratio (DLR) loss---%
have respectively been used as surrogate losses
in~\cite{carlini17} and~\cite{croce20aa}.

Our surrogate loss~\eqref{eq:surrogate}
has twofold advantages.
First, it prevents the gradients from
floating-point underflows
and improves convergence.
Second, unlike the DL or the DLR loss,
it can still
represent the original SCE loss
in a faithful fashion,
and all logits
can still contribute to the final loss.

Finally, we define a targeted variant
of \( \surrogateloss \),
which moves the logits
towards a predefined target class \( k \),
with a one-hot vector \( \target = \onehot(k, K) \):
\begin{equation}
    \targetloss(\z, \y) \triangleq -\sceloss\!\left(
        \frac\z{t} {\left(
            \y^\top \z -
            \max\left(
                \left( 1 - \y \right) \cdot \z
            \right)
        \right)}^{-1}, \target
    \right),
    \label{eq:surrogate:target}
\end{equation}

\subsection{Summary}\label{sec:method:summary}

In addition
to the original contributions
explained above,
we also employ simple yet helpful tactics
from previous literatures.
First, a step-size schedule
with a linear decay
\( 2 \epsilon (1 - i/I), \)
is used in the iterative updates,
where \( \epsilon \)
is the \textlnorm{\infty}
perturbation boundary \( \epsilon \)
in \eqref{eq:adversarial},
\( i \) is the current iteration number
and \( I \) denotes the total number of iterations.
Second, we adapt momentum-based updates
from~\cite{croce20aa}.

To summarize,
we illustrate the \attack{} algorithm
in \Cref{alg:overview}.
The function \(\mathtt{\attack\_\,Attack}\)
accepts the following inputs:
the model \( f_\allw \),
the pretrained logits function
for the \ordinal{\(l\)} layer
to be attacked jointly,
natural image \( \x \)
and its one-hot label \( \y \),
the step-size \( \stepsize \),
the interpolation parameter \( \beta \)
between the \ordinal{\(l\)} layer
and the output layer,
the momentum weight used \( \momentum \),
the perturbation boundary \( \epsilon \),
and lastly the maximum iteration count \( I \).

With the algorithm above,
we can perform a simple grid search
on \( \beta \in [0, 1] \).
As using latent features
results in faster convergence,
the search can start
from a point in the middle (\eg~0.5)
to minimize the number of iterations
required to attack each image.
Finally, to further push the limit
of \attack{},
we could incorporate
the \emph{multi-targeted}
attack~\cite{gowal19surrogate},
\ie~after
the untargeted surrogate loss~\eqref{eq:surrogate},
we enumerate
\( k \in [1:K]\,/\arg\max\y \),
\ie~all possible target classes
except for the ground truth,
with the targeted surrogate loss
\( \targetloss \)
defined in~\eqref{eq:surrogate:target}.
For computational efficiency,
the above search procedure
can be early-stopped
for successful adversarial examples,
until all images in \(\attackset\)
have been sift through progressively.

\section{Experiments}\label{sec:results}

For a fair comparison
against existing work
on adversarial attacks
across different defense techniques,
in our evaluation
we used the most common
\textlnorm{\infty} threat model
on the \cifarx{}
and \cifarc{} datasets~\cite{cifar}.
We extracted
from recent publications
with open PyTorch defense models for testing,
and ensured the list to be
as comprehensive as possible.
We reproduced traditional white-box attacks
(PGD~\cite{madry18},
 BIM~\cite{kurakin17adversarial},
 MIM~\cite{dong18momentum},
 FAB~\cite{croce20fab}),
all with 100 iterations
and a constant step-size of \( 2/255 \)
as baselines.
We use \att{MT}{\( I, B \)}
to denote flavors of \attack,
where \( I \)
is the number of gradient iterations used,
\( B \) is the number of \( \beta \) values searched,
and \texttt{MT} denotes the use of multi-targets.
The \cifarx{} testing set
is assumed to be \( \attackset \),
and the momentum of the iterative updates
is \( \momentum = 0.75 \).
A full comparison
can be found in~\Cref{tab:compare}.
It reveals that
not only is \att{MT}{100,6}
an even stronger attack
than \emph{AutoAttack} (AA)~\cite{croce20aa},
the current SOTA
in the robustness evaluation
of defense strategies\footnote{%
    The column ``AA'' was retrieved
    from \url{https://github.com/fra31/auto-attack}
    as of November 15, 2020.
},
but also it has
a better worst-case complexity
in terms of the maximum number
of forward-passes
required for each image.
To push the boundary of \attack,
we also report \att{MT}{1\unitk,10},
which enjoys a substantial increase
in compute effort.
Note that
AA as an ensemble
combines multiple strategies
(PGD with momentum
 and two surrogate losses~\cite{croce20aa},
 square attack~\cite{andriushchenko20square}
 and FAB~\cite{croce20fab}),
whereas \attack{}
uses only~\Cref{alg:overview} uniformly.
\begin{algorithm}[t]
\caption{%
    The \attack{} white-box attack.}\label{alg:overview}
\algnewcommand{\IfThen}[2]{
    \State \algorithmicif\ {#1}\ \algorithmicthen\ {#2}}
\algnewcommand{\IfThenElse}[3]{
    \State \algorithmicif\ {#1}\ %
    \algorithmicthen\ {#2}\ \algorithmicelse\ {#3}}
\newcommand{\algcmt}{\algorithmiccomment}
\begin{algorithmic}[1]
    \Function{\tt\attack\_\,Attack}{$
        f_\allw, \logit{l},
        \x, \y,
        \stepsize, \beta, \momentum,
        \epsilon, I
    $}
        \State{$
            \xadv_0 \gets \project\left(
                \x + \mathbf{u}
            \right) $,
        }
        \algcmt{Optional random start}
        \State{
            \hphantom{$\xadv_0 \gets \!$}
            where $
            \mathbf{u} \sim \uniform{[-\epsilon, \epsilon]}
        $}
        \State{$\m_0 \gets 0$}
        \For{$
            i \in [0:I-1]
        $}
            \State{$
                \z_o \gets f_\allw(\xadv_i)
            $,\, $
                \z_l \gets \logit{l}(\f{l}(\xadv_i))
            $}
            \State{$
                \sigma_o
                \gets
                    \y^\top \z_o -
                    \max\left(
                        \left( 1 - \y \right) \cdot \z_o
                    \right)
            $}
            \IfThen{$\sigma_o \leq 0$}{\Return{$\xadv_i$}}
            \algcmt{Successful attack}
            \State{$
                \sigma_l \gets
                    \y^\top \z_l -
                    \max\left(
                        \left( 1 - \y \right) \cdot \z_l
                    \right)
            $}
            \State{%
                $\beta_i \gets 1$
                \algorithmicif{} $\sigma_l \leq 0$
                \algorithmicelse{} $\beta$
            }
            \algcmt{Adaptive weight}
            \State{$
                \z \gets
                    \beta_i \frac{\z_o}{\sigma_o} +
                    (1 - \beta_i) \frac{\z_l}{\sigma_l}
            $}
            \algcmt{Surrogate loss}
            \State{$
                \bm{g}_{i + 1} \gets
                    \sign\left(\nabla_{\xadv_i} \!
                        \sceloss\left( \frac\z{t}, \y \right)
                    \right)
            $}
            \State{$
                \stepsize \gets 2 \epsilon (1 - \frac{i}{I})
            $}
            \algcmt{Linear step-size schedule}
            \State{$
                \m_{i + 1} \gets \project\left(
                    \m_i + \stepsize \bm{g}_{i + 1}
                \right)
            $}
            \algcmt{Momentum}
            \State{$
                \xadv_{i + 1} \gets \project\left(
                    \xadv_i + {}
                \right.
                $}
            \algcmt{Iterative update}
            \State{$
                \left.
                    \quad
                    \momentum \left( \m_{i+1} - \xadv_i \right) +
                    (1 - \momentum) \left( \xadv_i - \xadv_{i - 1} \right)
                \right)
            $}
        \EndFor{}
        \State{\Return{$\xadv_I$}}
        \algcmt{Give up after $I$ iterations}
    \EndFunction%
\end{algorithmic}
\end{algorithm}%
\begin{table*}[ht]
\centering%
\newcommand{\tcenterhack}[1]{\multicolumn{1}{c|}{#1}}
\newcommand{\tnahack}{\tcenterhack{---}}
\newcommand{\aaa}{\textbf{AA}\textsubscript{100}}
\caption{%
    Comparing accuracy under attack (\%)
    of \attack{}
    against iterative methods~\cite{
        madry18,kurakin17adversarial,
        dong18momentum,croce20fab}
    and AutoAttack (\aaa)~\cite{croce20aa}
    across various defense strategies.
    The ``\( \bm\Delta \)'' column shows the difference
    between the reported (``\textbf{Nominal}'')
    and \attack{} accuracies.
    Models marked with \textdagger{}
    were additionally trained
    with unlabeled datasets~\cite{80mti}.
    We used \( \epsilon = 8/255 \)
    except for models marked with \textdaggerdbl,
    which used \( \epsilon = 0.031 \)
    as originally reported by the authors.
}\label{tab:compare}
\adjustbox{max width=\textwidth}{%
\begin{tabular}{l|LL|LLLLL|LLL}  
\toprule
\thead{\cifarx{} defense method}
    & \thead{Clean} & \thead{Nominal}
    & \thead{PGD} & \thead{MIM} & \thead{BIM} & \thead{FAB}
    & \tcenter{\aaa}
    & \tcenter{\attmt{100,6}} & \tcenter{\attmt{1\unitk,10}}
    & \thead{\(\bm\Delta\)} \\
\tcenter{Worst-case complexity}
    & \tcenter{1} & \tcenter{}
    & \tcenter{100} & \tcenter{100} & \tcenter{100} & \tcenter{100} & \tcenter{8.3\unitk}
    & \tcenter{6\unitk} & \tcenter{100\unitk} & \\
\midrule
Adversarial weight perturbation~\cite{wu20wp}\supdagger{}
    & 88.25 & 60.04
    & 63.26 & 66.17 & 65.33 & 60.72 & 60.04
    & 59.97 & \tbnum{59.94} & -0.10 \\
Unlabeled~\cite{carmon19}\supdagger{}
    & 89.69 & 62.5
    & 62.17 & 67.21 & 65.77 & 60.85 & 59.53
    & 59.42 & \tbnum{59.37} &  -3.13 \\
HYDRA~\cite{sehwag20hydra}\supdagger{}
    & 88.98 & 59.98
    & 59.98 & 65.29 & 63.87 & 58.33 & 57.14
    & 57.02 & \tbnum{56.98} &  -3.00 \\
Misclassification-aware~\cite{wang20misclass}\supdagger{}
    & 87.5  & 65.04
    & 62.55 & 66.74 & 65.57 & 57.60 & 56.29
    & 56.13 & \tbnum{56.07} &  -8.97 \\
Pre-training\cite{hendrycks19}\supdagger{}
    & 87.11 & 57.4
    & 57.54 & 61.51 & 60.25 & 55.58 & 54.92
    & 54.80 & \tbnum{54.74} &  -2.66 \\
Hypersphere embedding~\cite{pang20hypersphere}
    & 85.14 & 62.14
    & 62.17 & 63.66 & 63.15 & 54.47 & 53.74
    & 53.68 & \tbnum{53.64} &  -8.50 \\
Overfitting~\cite{rice20overfitting}
    & 85.34 & 58.0
    & 57.24 & 60.48 & 59.49 & 54.3  & 53.42
    & 53.34 & \tbnum{53.32} &  -4.68 \\
Self-adaptive training~\cite{huang20selfadaptive}\supddagger{}
    & 83.48 & 58.03
    & 56.58 & 60.07 & 58.89 & 54.52 & 53.33
    & \tbnum{53.19} & \tbnum{53.19} &  -4.84 \\
TRADES~\cite{zhang19trades}\supddagger{}
    & 84.92 & 56.43
    & 55.50 & 59.16 & 58.06 & 53.96 & 53.08
    & 52.98 & \tbnum{52.94} &  -3.49 \\
Robustness (Python Library)~\cite{engstrom19}
    & 87.03 & 53.29
    & 52.32 & 58.39 & 56.52 & 50.67 & 49.21
    & 49.11 & \tbnum{49.09} & -4.20 \\
YOPO~\cite{zhang19yopo}
    & 87.20 & 47.98
    & 46.15 & 52.57 & 50.69 & 45.80 & 44.83
    & 44.73 & \tbnum{44.71} & -1.44 \\
Fast adversarial training~\cite{wong20fast}
    & 83.8  & 46.44
    & 46.44 & 52.02 & 50.53 & 44.52 & 43.41
    & 43.29 & \tbnum{43.27} &  -3.17 \\
MMA training~\cite{ding20mma}
    & 83.28 & 47.18
    & 47.29 & 56.44 & 54.65 & 46.88 & 40.21
    & 39.84 & \tbnum{39.74} & -7.44 \\
Neural level sets~\cite{atzmon19levelsets}\supddagger{}
    & 81.3  & 79.67
    & 79.83 & 79.83 & 79.83 & 40.98 & 40.22
    & 39.81 & \tbnum{39.77} & -39.90 \\
Feature scattering~\cite{zhang19scatter}
    & 89.98 & 60.6
    & 69.01 & 75.66 & 74.54 & 43.42 & 36.64
    & 36.02 & \tbnum{35.94} & -24.66 \\
Adversarial interpolation~\cite{zhang20interpolation}
    & 90.25 & 68.7
    & 73.13 & 75.84 & 74.95 & 43.34 & 36.45
    & 35.24 & \tbnum{35.14} & -33.56 \\
Sensible adversarial training~\cite{kim20sensible}
    & 91.51 & 57.23
    & 59.93 & 68.79 & 66.85 & 41.87 & 34.22
    & 33.39 & \tbnum{33.33} & -23.90 \\
Stylized adversarial defense~\cite{naseer20stylized}
    & 93.29 & 78.68
    & 78.68 & 82.87 & 82.50 & 19.14 & 13.42
    & 11.16 & \tbnum{11.09} & -67.59 \\
Manifold regularization~\cite{jin20manifold}
    & 90.84 & 77.68
    & 77.68 & 77.63 & 77.30 & 27.18 &  1.35
    & 0.22 & \tbnum{0.21}  & -77.47 \\
Polytope conformity loss~\cite{mustafa19pcl}
    & 89.16 & 32.32
    & 19.42 & 42.88 & 38.77 &  5.81 &  0.28
    & 0.06 & \tbnum{0.03}  & -32.29 \\
\bottomrule
\toprule
\thead{\cifarc{} defense method}
    & \thead{Clean} & \thead{Nominal}
    & \thead{PGD} & \thead{MIM} & \thead{BIM} & \thead{FAB}
    & \tcenter{\aaa}
    & \tcenter{\attmt{100,6}}
    & \tcenter{\attmt{1\unitk,10}}
    & \thead{\(\bm\Delta\)} \\
\tcenter{Worst-case complexity}
    & \tcenter{1} & \tcenter{}
    & \tcenter{100} & \tcenter{100} & \tcenter{100} & \tcenter{100} & \tcenter{8.3\unitk}
    & \tcenter{6\unitk} & \tcenter{100\unitk} & \\
\midrule
Adversarial weight perturbation~\cite{wu20wp}
    & 60.38 & 28.86
    & 33.68 & 35.24 & 34.70 & 29.25 & 28.86
    & 28.79 & \tbnum{28.77} & -0.09 \\
Pre-training~\cite{hendrycks19}\supdagger{}
    & 59.23 & 33.5
    & 33.70 & 35.66 & 35.09& 28.83 & 28.42
    & 28.25 & \tbnum{28.23} & -5.27 \\
Progressive Hardening~\cite{sitawarin20ates}
    & 62.82 & 24.57
    & 26.75 & 30.78 & 29.66 & 25.14 & 24.57
    & 24.50 & \tbnum{24.48} & -0.09 \\
Overfitting~\cite{rice20overfitting}
    & 53.83 & 28.1
    & 20.95 & 24.0 & 23.10 & 19.49 & 18.95
    & 18.87 & \tbnum{18.86} & -9.24 \\
\bottomrule
\end{tabular}
}
\end{table*}%

\begin{figure}[ht]
    \includegraphics[
        width=\linewidth,
        trim=0 15pt 0 0
    ]{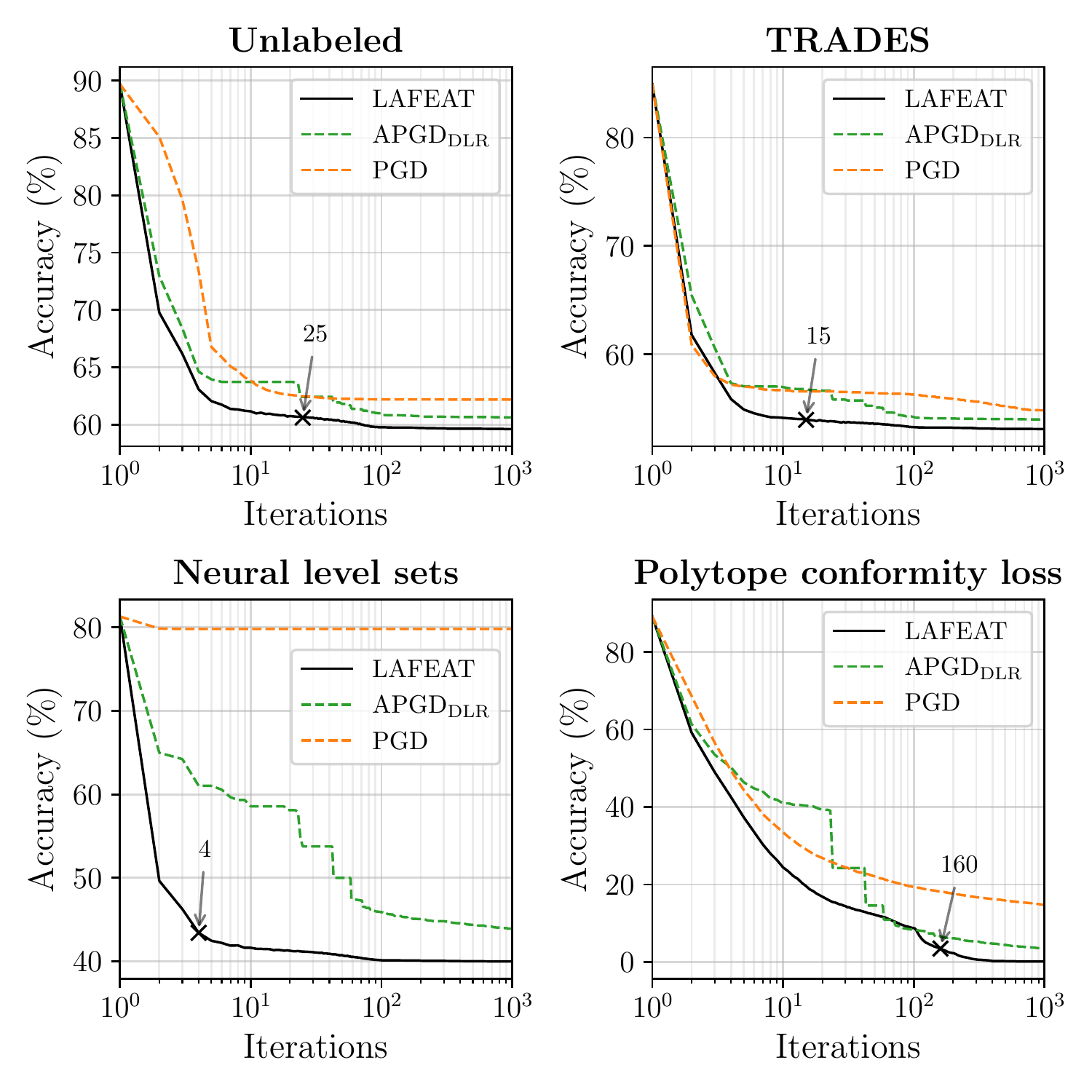}
    \caption{%
        Comparing the performance
        of \att{}{100,10}
        against other adversarial attack methods
        (\apgddlr~\cite{croce20aa}
         and PGD~\cite{madry18})
        on defenders~\cite{
            carmon19,zhang19trades,
            atzmon19levelsets,mustafa19pcl}.
        The horizontal and vertical axes
        respectively show
        the number of iterations used so far,
        and the percentage
        of remaining unsuccessful examples.
        The iteration count
        needed for \attack{}
        to defeat \apgddlr-100 (10 restarts)
        is also marked.
    }\label{fig:results:cost}
\end{figure}

\textbf{Faster convergence.}
It is sensible to argue that
models could also rely
on \emph{computational security}
as one of their defense tactics.
We would like to highlight
that in contrast to most attack methods,
the effectiveness of \attack{}
is not accompanied by high computational costs.
We found that
by exploiting latent features,
it generally leads
to faster convergences
to adversarial examples
than competing methods.
In~\Cref{fig:results:cost},
we compare the speed of convergence
among three different methods.
For baseline, we used PGD-1000.
We picked \apgddlr{}
with 100 iterations and 10 restarts,
the most effective attack method
of the AA ensemble~\cite{croce20aa}
across most defense methods in~\Cref{tab:compare}.
For computational fairness,
we selected \att{}{100,10} to compete against it,
which is an untargeted \attack{}
with a 10-valued
\( \beta \in \{ 0.0, 0.1, 0.2, \ldots, 0.9 \} \)
grid search starting from \( 0.5 \),
where each value used only 100 iterations.
The results show that
for all 4 defending models,
\attack{}
is not only stronger,
but also often orders of magnitude
faster than \apgddlr{} and PGD
for successful attacks.
Finally,
the logits layers
introduce minuscule overhead
(\( \leq\!\!{0.008\%} \) in all models),
and have no discernible impact
on the iteration time.%

\textbf{%
    Adversarial-trained latent features
    improve model robustness.}
As demonstrated earlier
with \attack,
one can exploit the latent features
learned by defending models
to craft powerful adversarial examples.
A question then ensues:
\emph{%
    is it possible
    to fortify latent features against attacks
    to improve model robustness?
}
To answer this,
we carried out a simple experiment
and trained two \wrn-32-10 models,
both with ordinary
PGD-7 adversarial training from~\cite{madry18}.
The only difference is that
in one of them
we additionally introduced
logits layers for residual block outputs
to be adversarial trained
along with the output layer.
For attacks,
we likewise ablated
with the use of latent features.
The results can be found in \Cref{tab:training}.
Note that
the model with robust latent features
displayed a better defense against all attacks
than the other.
Even when faced against attacks
that do no leverage latent features,
Perhaps the most revealing result is:
training latent features
to be more robust
can improve the overall robustness
even when faced against attacks
without using latent features.%
\begin{table}[t]
\centering%
\caption{%
    PGD-7 adversarial training \vs{}
    \att{}{100,10},
    both explored with (+LF) and without (--LF) latent features.
}\label{tab:training}%
\begin{tabular}{@{}rr|c|ccc@{}}
\toprule
\multicolumn{2}{c}{\textbf{Accuracy under}}
    &
    & \multicolumn{3}{c}{\textbf{Adversarial Attack}} \\
\multicolumn{2}{c}{\textbf{attack (\%)}}
    & \text{Clean}
    & \text{PGD-100}
    & \text{--LF}
    & \text{+LF} \\
\midrule
\multirow{2}{*}{\textbf{Defense}}
    & \text{--LF}
    & 79.56
    & 47.06
    & 41.21
    & 41.05 \\
    & \text{+LF}
    & 83.53
    & 47.17
    & 41.50
    & 41.26 \\
\bottomrule%
\end{tabular}%
\vspace{-8pt}%
\end{table}%

\textbf{Compression \vs{} robustness.}
In \Cref{tab:compress},
compressed models from HYDRA~\cite{sehwag20hydra}
displayed a worsening robustness degradation
between the reported (PGD-50 with 10 restarts)
and \att{}{100,10}
for an increasing pruning ratio.
This shows
that model-holistic attacks
can potentially overestimate
the robustness of compressed models.%
\begin{table}[ht]
\centering
\caption{%
    The effect of \att{}{100}
    on the adversarial trained
    and compressed \wrn-28-4 models
    from HYDRA~\cite{sehwag20hydra}.
    PR denotes pruning ratio,
    \ie~the percentage of zeros
    in model weights.
}\label{tab:compress}
\begin{tabular}{@{}r|LLLL@{}}
\toprule
\thead{PR (\%)}
    & \thead{Clean}
    & \thead{Nominal}
    & \thead{\attack}
    & \thead{\( \bm\Delta \)} \\
\midrule
     0\% & 85.6 & 57.2 & 53.94 & -3.26 \\
    90\% & 83.7 & 55.2 & 51.68 & -3.52 \\
    95\% & 82.7 & 54.2 & 49.87 & -4.33 \\
    99\% & 75.6 & 47.3 & 42.73 & -4.57 \\
\bottomrule
\end{tabular}
\vspace{-8pt}%
\end{table}
\begin{figure}[ht]
    \centering
    \includegraphics[width=\linewidth, trim=0 15pt 0 15pt]{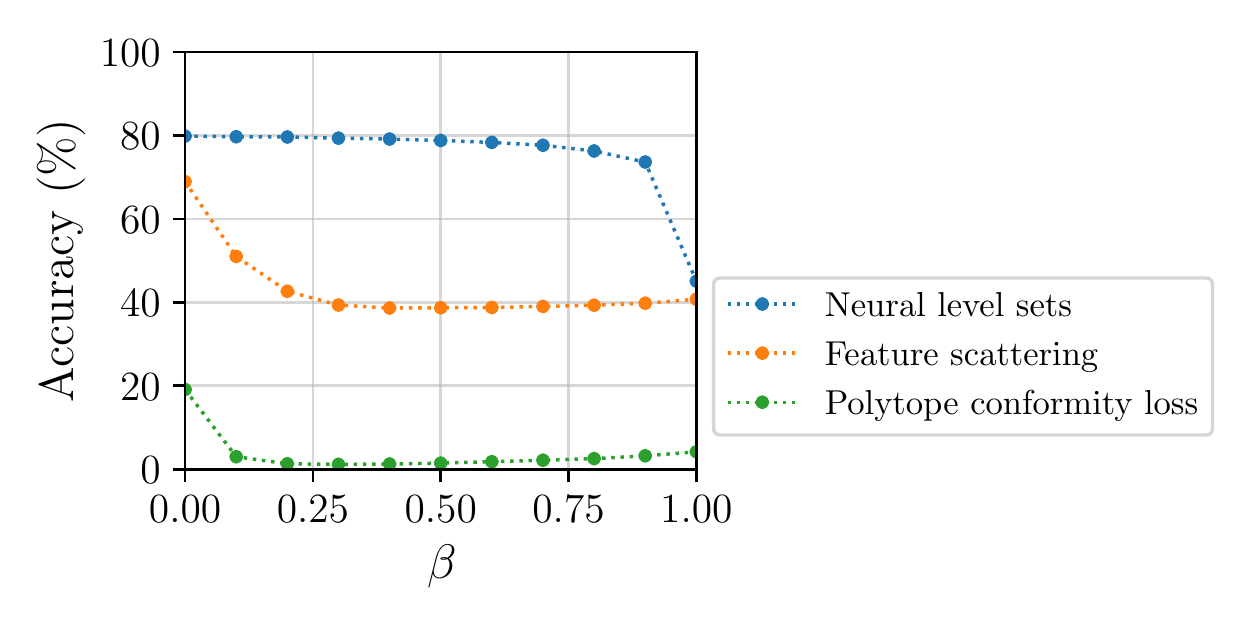}
    \caption{%
        Accuracy under attack \vs{} \( \beta \),
        the interpolation
        between the output and latent logits
        for \cite{
            atzmon19levelsets,zhang19scatter,
            mustafa19pcl},
        Here, \( \beta = 0 \) uses only the output logits,
        and \( 1 \) uses only the latent logits.
    }\label{fig:results:beta}
    \vspace{-8pt}
\end{figure}

\textbf{Interpolation between latent and output logits.}
\Cref{fig:results:beta}
varies the weight
between the most effective latent feature
and the final logits output
for \att{}{100,10}.
It used \( \beta \in \{ 0.0, 0.1, \ldots, 0.9 \} \)
and 100 iterations for each.
The result showed that
different defense strategies
call for distinct \( \beta \) values,
making the search of \( \beta \)
a compelling necessity.

\begin{figure}[ht]
    \includegraphics[width=\linewidth]{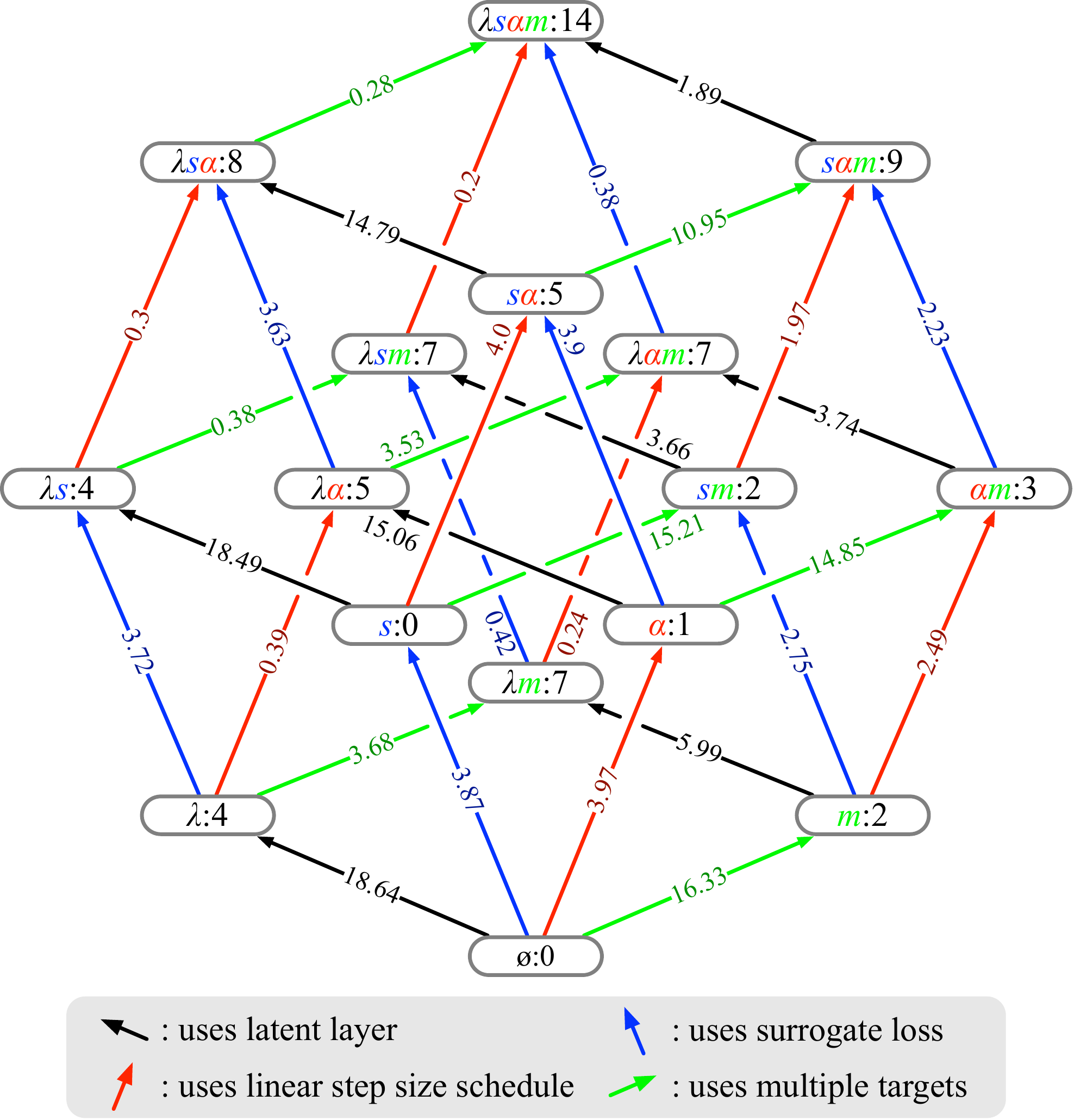}
    \caption{%
        We ablate the full \attmt{100}
        of its 4 tactics
        on a complete lattice.
        Here, each node
        is a unique combination
        of 4 tactics,
        where
        \( \color{black}{\lambda} \) uses
        a latent layer
        with \( \beta \) search;
        \( \color{blue}{s} \) is
        the surrogate loss instead
        of the standard SCE loss;
        \( \color{red}{\alpha} \) uses
        the linear step-size schedule;
        and \( \color{green}{m} \)
        denotes multiple targets,
        and the number
        indicates the number of methods
        where \attack{}
        is better than AA
        across 14 different defenses~\cite{
            carmon19,sehwag20hydra,wang20misclass,
            hendrycks19,huang20sat,zhang19trades,
            wong20fast,atzmon19levelsets,zhang19scatter,
            zhang20interpolation,kim20sensible,
            jin20manifold,mustafa19pcl,naseer20stylized}.
        The arrows
        \inlinegraphics{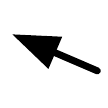},
        \inlinegraphics{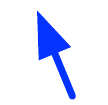},
        \inlinegraphics{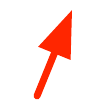} and
        \inlinegraphics{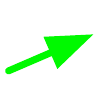}
        represent the introduction
        of the corresponding tactics,
        and each number on an arrow
        indicates the average accuracy degradation
        as a result of adding that tactic.
    }\label{fig:results:lattice}
    \vspace{-12pt}
\end{figure}

\textbf{Ablation analysis.}
\Cref{fig:results:lattice}
performs ablation analysis
of 4 tactics employed by \attack{}
across 14 defending models,
where 3 were introduced in this paper,
\ie~the use of latent features,
a new surrogate loss,
and a linear decay step-size schedule.
The last one is the multi-targeted attack
inspired by~\cite{gowal19surrogate}.
Across all 16 combinations of them,
we discovered that
adding latent features
to an existing combination of methods
always brings the greatest accuracy impact
among possible choices.

\section{Conclusion}\label{sec:conclusion}

\attack{} demonstrated
that exploiting latent features
is highly effective
against many recent defense techniques.
It efficiently outperformed
the current SOTA attack methods
across a wide-range of defenses.
We believe that the future progress
of adversarial attack and defense on CNNs
depends on the understanding
of how latent features
can be effectively used
as novel attack vectors.
The evaluation
of adversarial robustness,
therefore, cannot view the model
from a holistic perspective.
We made \attack{}
open-source\footnote{%
Available at: \url{https://github.com/lafeat/lafeat}.}
and hope it could pave the way
for gaining knowledge
on robustness evaluation
from the explicit use
of latent features.

\section*{Acknowledgments}

{\small
This work is supported in part by
National Key R\&D Program of China (\numero{2019YFB2102100}),
Science and Technology Development Fund of Macao S.A.R (FDCT)
under \numero{0015/2019/AKP},
Key-Area Research and Development Program
of Guangdong Province (\numero{2020B010164003}),
the National Natural Science Foundation of China
(\numeros{61806192, 61802387}),
Shenzhen Science and Technology Innovation Commission
(\numeros{JCYJ20190812160003719, JCYJ20180302145731531}),
Shenzhen Discipline Construction Project
for Urban Computing and Data Intelligence,
Shenzhen Engineering Research Center
for Beidou Positioning Service Technology
(\numero{XMHT20190101035}),
and Super Intelligent Computing Center,
University of Macau.
\par
}

\clearpage

{\small
\bibliographystyle{ieee_fullname}
\bibliography{references}
}

\end{document}